\ifdefined\pdfminorversion\pdfminorversion=7\relax\fi
\documentclass[preprint,12pt]{elsarticle}

\usepackage{amsmath,amssymb,amsfonts}
\usepackage{booktabs}
\usepackage{placeins}
\usepackage{float}
\usepackage{graphicx}
\usepackage[hidelinks]{hyperref}
\usepackage{setspace}

\journal{Computer Methods and Programs in Biomedicine}

\makeatletter
\def\ps@pprintTitle{%
  \let\@oddhead\@empty
  \let\@evenhead\@empty
  \let\@oddfoot\@empty
  \let\@evenfoot\@empty}
\renewcommand\paragraph{\@startsection{paragraph}{4}{\z@}%
  {-1.5ex \@plus -0.5ex \@minus -0.2ex}%
  {0.5ex \@plus 0.2ex}%
  {\normalfont\normalsize\bfseries}}
\makeatother

\newcommand{\E}{\mathbb{E}}
\newcommand{\abstractheading}[1]{%
    \par\nobreak\vspace*{0.5\baselineskip}%
    \noindent\textbf{#1}\par\nobreak\vspace*{0.4\baselineskip}\noindent}
\keywordtitlesep{\quad}

\protected\def\equalcontribmark{\textdagger}
\begin{document}
\doublespacing
\emergencystretch=2em
\clubpenalty=10000
\widowpenalty=10000
\displaywidowpenalty=10000

\begin{frontmatter}

\title{Fast Self-Supervised Generative CT Denoising via Few-Step Sampling}
\date{}

\renewcommand{\thefnote}{\equalcontribmark}
\pdfstringdefDisableCommands{\def\fnref#1{}\def\corref#1{}}
\author{Xingyi Yao\fnref{equal}}
\author{Zhengtao Ma\fnref{equal}}
\author{Zhennong Chen\corref{correspondence}}
\fntext[equal]{Xingyi Yao and Zhengtao Ma contributed equally to this work.}
\cortext[correspondence]{Corresponding author: Zhennong Chen (\href{mailto:Zhennong.Chen@xjtlu.edu.cn}{Zhennong.Chen@xjtlu.edu.cn}).}

\begin{abstract}
\abstractheading{Background and Objectives}
Low-dose and thin-slice computed tomography (CT) images suffer from elevated noise that obscures anatomical detail, while clean references for supervised denoising are often unavailable. Although Noise2Noise permits self-supervised training from noisy pairs, direct regression tends to oversmooth fine structure. Averaging multiple independently generated samples mitigates this blur, though repeating multi-step diffusion sampling is costly. We introduce Noise2Noise with Improved Mean Flow (N2N-iMF) to accelerate this process via few-step sampling.

\abstractheading{Methods}
N2N-iMF learns to generate one noisy image conditioned on its paired counterpart; at inference, independent outputs are sampled and averaged to suppress noise. We evaluated image quality and sampling efficiency on simulated thin-slice brain and low-dose abdominal CT across four sampling methods and averaging counts from 1 to 20, and validated the approach on real-world photon-counting CT (PCCT) brain scans.

\abstractheading{Results}
With ten samples per conditioning input and three network evaluations per sample, N2N-iMF achieved a mean absolute error of 2.60~HU on brain CT and 14.02~HU on abdominal CT, outperforming 50-evaluation denoising diffusion implicit model (DDIM) sampling (2.91 and 14.53~HU). On the same GPU, per-output generation time dropped from 7.29 to 0.43~s. Across eight PCCT cases, N2N-iMF reached a mean gray--white matter contrast-to-noise ratio of 1.341, versus 0.916 for Noise2Noise and 1.176 for DDIM.

\abstractheading{Conclusions}
N2N-iMF accelerates self-supervised generative CT denoising without distillation while keeping the number of averaged outputs adjustable at inference.
\par\vspace{12pt}
\end{abstract}

\begin{samepage}
\begin{keyword}
computed tomography \sep image denoising \sep self-supervised learning \sep flow-based generative
modeling \sep photon-counting CT \sep few-step inference
\end{keyword}
\end{samepage}

\end{frontmatter}
\raggedbottom

\section{Introduction}
\label{sec:introduction}

Low-dose computed tomography (CT) lowers patient radiation exposure, but reduced photon counts increase image noise and degrade the visibility of low-contrast anatomy~\cite{mccollough2017aapm}. A similar trade-off arises in thin-slice reconstruction, where improved through-plane resolution comes at the cost of elevated noise~\cite{chen2026thin}, as well as in high-resolution photon-counting CT (PCCT)~\cite{willemink2018pcct,hein2024pcct}. Although deep networks can suppress CT noise effectively, supervised training relies on spatially aligned low- and normal-dose image pairs. Unpaired methods relax the requirement for pixel-level alignment, yet they still need a separate set of high-quality scans that may be difficult to obtain.

Noise2Noise eliminates the need for clean training targets by learning directly from paired noisy observations~\cite{lehtinen2018n2n}. When two observations share the same underlying anatomy and carry independent, zero-mean noise, predicting one from the other has the same squared-error optimum as supervised training with clean images. In CT, suitable pairs can be formed either by reconstructing disjoint subsets of projection data or by pairing adjacent thin slices with nearly identical anatomy~\cite{wu2019consensus,chen2026thin}. Because squared-error regression estimates the conditional mean of the clean image, however, the denoised output tends to oversmooth fine anatomical detail.

To preserve fine structure without clean references, Chen et al.~\cite{chen2026thin} trained a conditional diffusion model to learn the distribution of noisy targets and averaged multiple independent samples at inference. Although denoising diffusion implicit models (DDIM) reduce the number of steps required for each output~\cite{song2021ddim}, generating multiple realizations still scales the total inference cost with the number of outputs. In contrast, Mean Flow and Improved Mean Flow (iMF) predict the velocity averaged over a finite time interval rather than the instantaneous velocity, allowing larger updates along the trajectory and generation in only one or a few network evaluations~\cite{geng2025meanflow,geng2026imf}.

In this work, we introduce Noise2Noise with Improved Mean Flow (N2N-iMF) to accelerate sample-averaged generative CT denoising. Our main contributions are as follows:
\begin{enumerate}
\item We formulate N2N-iMF, a self-supervised few-step conditional generator trained solely on paired noisy images without clean references, which suppresses noise by averaging independent generated outputs while retaining fine anatomical structure.
\item Using three network evaluations per output ($N_{\mathrm{FE}}=3$), N2N-iMF generates each output approximately 17 times faster than DDIM with 50 evaluations, making averaging over $K$ independent realizations practical without distillation while keeping $K$ adjustable at inference.
\item We validate N2N-iMF on simulated low-dose abdominal CT, simulated thin-slice brain CT, and real-world PCCT, demonstrating improved noise suppression and detail preservation over competing self-supervised methods in both quantitative metrics and visual assessments.
\end{enumerate}

\section{Related Work}
\label{sec:related_work}

\subsection{Supervised and unpaired CT denoising}
Supervised CT denoisers, such as RED-CNN~\cite{chen2017redcnn} and the directional wavelet network of Kang et al.~\cite{kang2017wavelet}, learn direct mappings from low-dose inputs to aligned higher-quality targets. Because pixelwise losses tend to oversmooth fine texture, later approaches incorporated adversarial and perceptual objectives to better preserve image sharpness~\cite{yang2018wganvgg,wolterink2017gan}. When aligned pairs are unavailable, unpaired methods instead learn from separate low- and routine-dose collections, typically under a cycle-consistency constraint~\cite{zhu2017cyclegan,li2021unpaired}.

\subsection{Self-supervised CT denoising}
Self-supervised methods avoid clean reference scans altogether. Building on Noise2Noise, which replaces clean targets with independent noisy observations~\cite{lehtinen2018n2n}, CT denoising approaches construct training pairs from disjoint measurement subsets or neighboring slices~\cite{wu2019consensus,zhang2021n2context,zhou2024neighboring}. When paired data are unavailable, blind-spot methods such as Noise2Void and Noise2Self predict held-out pixels from their surroundings~\cite{krull2019n2v,batson2019n2s}, whereas Noise2Sim pairs structurally similar regions as training targets~\cite{niu2023n2sim}. Taking a statistical approach, Noise2Score derives a clean estimate from the score of the noisy image distribution together with a noise model~\cite{kim2021n2score}.

More recently, generative self-supervised methods have been explored to mitigate regression blur. DDM$^2$ first applies a preliminary denoiser to estimate the noise level and matches the noisy input to an intermediate diffusion state before reverse denoising~\cite{xiang2023ddm2}. When paired noisy CT scans are available, Chen et al.~\cite{chen2026thin} instead train a conditional diffusion model to generate noisy targets and average independent outputs to obtain the denoised image.

\subsection{Flow-based generative models in medical imaging}
Flow matching and rectified flow learn a continuous velocity field that transports noise into images~\cite{lipman2023fm,liu2023rectified}. In medical imaging, flow-based modeling has been applied to MRI reconstruction using only undersampled measurements~\cite{luo2026gtf2m}. For CT, the Poisson Flow Generative Model++ (PFGM++) supplies an image prior for denoising~\cite{hein2024pcct}, and conditional Mean Flow has been studied across images reconstructed with different kernels and slice thicknesses~\cite{apurba2026racmf}.

Standard diffusion models generate images by iteratively predicting and removing added noise~\cite{ho2020ddpm}, which makes inference slow. Efforts to accelerate sampling generally follow two directions: reducing the number of numerical updates or learning a direct mapping to the trajectory endpoint. Along the first direction, DDIM reduces the required sampling steps~\cite{song2021ddim}, while elucidated diffusion models (EDM) improve both training and numerical solvers~\cite{karras2022edm}. Along the second, consistency models map noisy states directly to the endpoint~\cite{song2023consistency}, an approach adapted via consistency distillation in Poisson Flow Consistency Models for single-step CT denoising~\cite{hein2025pfcm}.

Rather than relying on fine time discretization or distillation, Mean Flow predicts the velocity averaged over a time interval so that sampling can advance in larger updates~\cite{geng2025meanflow}. iMF further reformulates its training as regression on the velocity at the current time~\cite{geng2026imf}. In N2N-iMF, we adopt this few-step generator to learn from noisy CT pairs and efficiently produce independent realizations for denoising by sample averaging.

\section{Methods}
\label{sec:methods}

\subsection{Supervised denoising model}
\label{sec:supervised_model}

A supervised generative denoiser learns to sample a clean CT image $z$ given its noisy observation
$y$, where $n$ denotes CT noise:
\begin{equation}
    p(z\mid y),\qquad y=z+n.
    \label{eq:supervised_distribution}
\end{equation}
Training such a model requires a paired clean target $z$ for each noisy input $y$, which is often unavailable in practice.

\subsection{Noise2Noise with Improved Mean Flow}
\label{sec:n2n_imf}

In the absence of clean targets, N2N-iMF is trained on paired noisy observations $y_1$ and $y_2$
that share the same underlying clean image $z$ but carry independent CT noise realizations $n_1$ and $n_2$.
Specifically, the conditional iMF generator learns to sample the noisy target $y_2$ given its paired input $y_1$:
\begin{equation}
    p(y_2\mid y_1),\qquad y_1=z+n_1,\quad y_2=z+n_2.
    \label{eq:n2n_distribution}
\end{equation}

\subsection{Denoising by sample averaging}
\label{sec:main_framework}

Because the clean image $z$ is unobserved during training, marginalizing over $z$ yields
\begin{equation}
\begin{aligned}
    p(y_2\mid y_1)
    &=\int p(y_2,z\mid y_1)\,\mathrm{d}z \\
    &=\int p(y_2\mid z,y_1)p(z\mid y_1)\,\mathrm{d}z.
\end{aligned}
    \label{eq:latent_decomposition}
\end{equation}
Here, $p(z\mid y_1)$ is the posterior clean-image distribution targeted by supervised denoising.
Because $n_1$ and $n_2$ are independent given $z$, the target $y_2$ is conditionally independent of $y_1$ once $z$ is fixed;
hence $p(y_2\mid z,y_1)=p(y_2\mid z)$, and Eq.~\eqref{eq:latent_decomposition} simplifies to
\begin{equation}
    p(y_2\mid y_1)=\int p(y_2\mid z)p(z\mid y_1)\,\mathrm{d}z.
    \label{eq:conditional_mixture}
\end{equation}
Drawing a realization from $p(y_2\mid y_1)$ is therefore equivalent to sampling a clean image from $p(z\mid y_1)$
and corrupting it with CT noise, so the $k$-th generated output can be expressed as
\begin{equation}
\begin{aligned}
    y_2^{(k)}&=z^{(k)}+n_2^{(k)},\\
    z^{(k)}&\sim p(z\mid y_1),\qquad
    n_2^{(k)}\sim p(n_2\mid z^{(k)}).
\end{aligned}
    \label{eq:conditional_sample}
\end{equation}
In practice, the network generates $y_2^{(k)}$ directly, with $n_2^{(k)}$ representing the CT noise associated with the latent clean image $z^{(k)}$. Averaging $K$ independent outputs gives
\begin{equation}
    \widehat{z}_K
    =\frac{1}{K}\sum_{k=1}^{K}y_2^{(k)}
    =\frac{1}{K}\sum_{k=1}^{K}z^{(k)}
     +\frac{1}{K}\sum_{k=1}^{K}n_2^{(k)}.
    \label{eq:sample_mean}
\end{equation}
Provided that the CT noise has zero conditional mean given $z$ and finite variance, averaging $K$ realizations reduces the noise standard deviation by a factor of $\sqrt{K}$. As $K$ grows, $\widehat{z}_K$ approaches $\E[z\mid y_1]$, the posterior mean targeted by squared-error regression. Unlike direct regression, however, modeling $p(y_2\mid y_1)$ provides access to individual realizations around this mean; averaging a finite number suppresses noise while retaining part of that variation, allowing $K$ to be selected at inference to balance denoising strength against image texture.

Whereas Chen et al.~\cite{chen2026thin} sampled from $p(y_2\mid y_1)$ using conditional diffusion, the same averaging framework applies to flow-based generators such as flow matching~\cite{lipman2023fm} and iMF~\cite{geng2026imf}. Let $N_{\mathrm{FE}}$ denote the number of network evaluations (NFE) required per generated output; averaging $K$ independent outputs then costs $K N_{\mathrm{FE}}$ evaluations for each conditioned input. By adopting iMF to keep $N_{\mathrm{FE}}$ small, N2N-iMF substantially reduces the computational overhead of repeated generation, and both $N_{\mathrm{FE}}$ and $K$ can be adjusted at inference without retraining.

\subsection{Training}
\label{sec:training}

The generator maps Gaussian noise to the noisy target $y_2$ conditioned on the paired image $y_1$. For each training pair $(y_1,y_2)$, we draw Gaussian noise $\epsilon\sim\mathcal{N}(0,\mathrm{Id})$ and two time points $0\leq s\leq t\leq1$, forming an intermediate state
\begin{equation}
    x_t=(1-t)y_2+t\epsilon.
    \label{eq:conditional_path}
\end{equation}
This linear path interpolates between the noisy target $y_2$ at $t=0$ and Gaussian noise $\epsilon$ at $t=1$, yielding a known velocity target $d=\epsilon-y_2$ without requiring a clean CT image. Note that $\epsilon$ is auxiliary noise sampled for generative training, distinct from the physical CT noise $n_2$ already present in $y_2$.

The conditional U-Net $u_\theta(x_t,y_1,s,t)$ predicts the average velocity over the interval $[s,t]$. During training, the Mean Flow identity converts this prediction into an instantaneous velocity estimate at time $t$~\cite{geng2025meanflow,geng2026imf}:
\begin{equation}
    V_\theta=u_\theta+(t-s)\operatorname{sg}\!\left[\frac{\mathrm{D}u_\theta}{\mathrm{D}t}\right].
    \label{eq:imf_prediction}
\end{equation}
Here $\operatorname{sg}[\cdot]$ denotes the stop-gradient operator applied to the derivative correction. Along a velocity field $v$, the total derivative is $\mathrm{D}u_\theta/\mathrm{D}t=\partial_t u_\theta+(\partial_{x_t}u_\theta)v$ with $y_1$ and $s$ held fixed, where the second term captures how the predicted average velocity changes as $x_t$ moves along $v$. Written in this equivalent velocity-regression form, the original Mean Flow sets $v=d$, causing $V_\theta$ to depend on the regression target itself. In contrast, iMF substitutes the network's own estimate $v=u_\theta(x_t,y_1,t,t)$, so $d$ serves strictly as the regression target without entering the derivative correction~\cite{geng2025meanflow,geng2026imf}.

A second output head $v_\theta^{\mathrm{aux}}$, sharing the U-Net backbone, directly predicts $d$ during training. Both velocity predictions are supervised with pixelwise mean squared error $\ell$:
\begin{equation}
    \mathcal{L}_{\mathrm{iMF}}
    =\E\!\left[w\,\ell(V_\theta,d)
      +0.5\,w_{\mathrm{aux}}\,\ell(v_\theta^{\mathrm{aux}},d)\right],
    \label{eq:training_loss}
\end{equation}
where the adaptive weights $w=\operatorname{sg}[(\ell(V_\theta,d)+0.01)^{-1}]$ and $w_{\mathrm{aux}}=\operatorname{sg}[(\ell(v_\theta^{\mathrm{aux}},d)+0.01)^{-1}]$ are held fixed during backpropagation, and the expectation is taken over training pairs, Gaussian noise $\epsilon$, and sampled time points.

For brain CT and PCCT, we additionally fine-tune the trained generator with a PatchGAN adversarial loss to better match the local texture of noisy targets~\cite{isola2017patchgan}, with the discriminator comparing high-frequency residuals obtained by subtracting a $7\times7$ local average.

\subsection{Inference}
\label{sec:inference}

At inference, generation starts from Gaussian noise $x_1=\epsilon$ while holding the conditioning input $y_1$ fixed.
Stepping backward from $t=1$ to $t=0$, the network $u_\theta$ predicts the average velocity over each subinterval $[s,t]$, and the state is updated via
\begin{equation}
    x_s=x_t-(t-s)u_\theta(x_t,y_1,s,t),\qquad s<t.
    \label{eq:inference}
\end{equation}
After $N_{\mathrm{FE}}$ updates, the terminal state $x_0$ gives one generated output $y_2^{(k)}$. Repeating this procedure from $K$ independent Gaussian initializations and averaging the resulting outputs yields the denoised estimate $\widehat{z}_K$ in Eq.~\eqref{eq:sample_mean}.


\section{Experiments}
\label{sec:experiments}

\subsection{Dataset description}
The simulated abdominal CT dataset comprised 10 normal-dose cases from the AAPM Low Dose CT Grand Challenge~\cite{mccollough2017aapm}, split into seven cases for training and three for testing. Images with 2~mm slice thickness served as clean references, and low-dose scans were simulated by adding Gaussian noise in the projection domain followed by filtered backprojection (FBP) with a soft-tissue kernel. Each sinogram was split into two disjoint subsets to form a Noise2Noise pair~\cite{wu2019consensus}.

For simulated thin-slice brain CT, we used 100 retrospective scans, with 84 reserved for training and 16 for testing. Volumes with 5~mm slice thickness were interpolated to 0.625~mm to serve as clean references, and noise was simulated using the same projection-domain procedure as for abdominal CT. Training pairs were constructed from triplets of consecutive slices, taking the middle slice as the target $y_2$ and stacking its two neighbors channel-wise as the conditioning input $y_1$.

The real-world PCCT dataset consisted of 33 portable brain scans acquired at 120~kVp, 6--7~mA, and 0.707~mm slice thickness, with 25 cases used for development and eight for testing. Pairs followed the same three-slice arrangement used in simulated brain CT. Both the retrospective brain CT and PCCT collections received institutional review board approval. Across all three datasets, images had a matrix size of $512\times512$ pixels.

\subsection{Implementation details}
The conditional U-Net~\cite{ronneberger2015unet} was implemented in PyTorch and trained using Adam with a learning rate of $10^{-4}$ for 200 epochs on an NVIDIA A800 GPU. Inference runtime was measured separately on an NVIDIA GeForce RTX 4080 Laptop GPU with a batch size of 1, yielding a median generation time per $512\times512$ output of 0.43~s for N2N-iMF ($N_{\mathrm{FE}}=3$) versus 7.29~s for DDIM ($N_{\mathrm{FE}}=50$).

For the main comparisons, we set $N_{\mathrm{FE}}=3$ and $K=10$ on simulated CT, and $N_{\mathrm{FE}}=5$ and $K=20$ on PCCT. On abdominal CT, $K$ outputs were generated from each of the two split reconstructions and all $2K$ outputs were averaged, requiring $2K N_{\mathrm{FE}}$ evaluations per slice; on brain CT and PCCT, a single conditioning input formed from the two neighboring slices required $K N_{\mathrm{FE}}$ evaluations. To evaluate the effect of sample averaging, we fixed $N_{\mathrm{FE}}=3$ and varied $K$ from 1 to 20 in integer steps across the three abdominal CT and 16 brain CT test cases. In this $K$-sweep, the brain CT model was evaluated prior to adversarial fine-tuning.

\subsubsection*{\textit{Comparison with existing self-supervised methods}}
We compared N2N-iMF with Noise2Noise, Noise2Score, and DDM$^2$~\cite{lehtinen2018n2n,kim2021n2score,xiang2023ddm2} on the same test cohorts. To examine the impact of domain shift, the visual comparisons in Fig.~\ref{fig:simulated} also included a supervised diffusion model trained with a different noise model and reconstruction kernel.

\subsubsection*{\textit{Comparison of expedited sampling methods}}
Within the same conditional generative framework (Sec.~\ref{sec:main_framework}), we compared N2N-iMF against DDIM~\cite{song2021ddim}, EDM~\cite{karras2022edm}, and flow matching~\cite{lipman2023fm} for accelerated inference. Each method generated $K$ independent noisy targets conditioned on the paired input and averaged them for denoising (Fig.~\ref{fig:nfe}). Comparisons spanned $N_{\mathrm{FE}}=2,3,5,10,20,30,50$ at $K=10$. For PCCT, the diffusion reference used DDIM with 100 evaluations and eight averaged outputs.

\subsubsection*{\textit{Evaluation metrics}}
Simulated images were evaluated using mean absolute error (MAE), structural similarity index (SSIM)~\cite{wang2004ssim}, and learned perceptual image patch similarity (LPIPS)~\cite{zhang2018lpips}, computed over intensity windows of $[-160,240]$~HU for abdominal CT and $[0,100]$~HU for brain CT. Two-sided $t$-tests were performed at $N_{\mathrm{FE}}=3$ and $K=10$ to compare N2N-iMF with DDIM and EDM on each metric. Quantitative results are reported as mean $\pm$ standard deviation across test cases.

Because low-noise reference scans were unavailable for real-world PCCT, evaluation relied on visual inspection and gray--white matter contrast-to-noise ratio (CNR) measured within manually drawn regions of interest (ROIs). Display windows were matched across methods, and no histogram equalization was applied to PCCT images.

\section{Results}
\label{sec:results}

\subsection{Comparison with self-supervised methods}
Across both simulated datasets, N2N-iMF achieved the lowest MAE and LPIPS and the highest SSIM among all compared self-supervised methods (Table~\ref{tab:main}). On brain CT, it lowered MAE from 4.12~HU for Noise2Noise to 2.60~HU, increased SSIM from 0.693 to 0.766, and reduced LPIPS from 0.0990 to 0.0497. On abdominal CT, N2N-iMF similarly reached an MAE of 14.02~HU, an SSIM of 0.731, and an LPIPS of 0.0462, improving on the corresponding Noise2Noise values of 14.60~HU, 0.695, and 0.1170.

\begin{table}[H]
\centering
\singlespacing
\caption{Main results on brain CT ($n=16$) and abdominal CT ($n=3$). N2N-iMF uses $N_{\mathrm{FE}}=3$ and $K=10$. Values are mean $\pm$ SD; bold indicates the best mean in each dataset.}
\label{tab:main}
\vspace{6pt}
\begingroup
\small
\setlength{\tabcolsep}{7pt}
\renewcommand{\arraystretch}{1.18}
\begin{tabular}{lccc}
\toprule
Method & MAE (HU) $\downarrow$ & SSIM $\uparrow$ & LPIPS $\downarrow$ \\
\midrule
\multicolumn{4}{l}{\textit{Brain CT}} \\
FBP & $6.28 \pm 0.59$ & $0.412 \pm 0.053$ & $0.1540 \pm 0.0240$ \\
Noise2Score & $4.91 \pm 0.68$ & $0.655 \pm 0.042$ & $0.0930 \pm 0.0090$ \\
Noise2Noise & $4.12 \pm 0.40$ & $0.693 \pm 0.032$ & $0.0990 \pm 0.0150$ \\
DDM$^2$ & $3.50 \pm 0.44$ & $0.733 \pm 0.035$ & $0.0680 \pm 0.0180$ \\
N2N-iMF & $\mathbf{2.60 \pm 0.36}$ & $\mathbf{0.766 \pm 0.035}$ & $\mathbf{0.0497 \pm 0.0131}$ \\
\midrule
\multicolumn{4}{l}{\textit{Abdominal CT}} \\
FBP & $23.80 \pm 4.50$ & $0.607 \pm 0.064$ & $0.1440 \pm 0.0420$ \\
Noise2Score & $16.20 \pm 2.90$ & $0.659 \pm 0.052$ & $0.1230 \pm 0.0430$ \\
Noise2Noise & $14.60 \pm 1.40$ & $0.695 \pm 0.026$ & $0.1170 \pm 0.0200$ \\
DDM$^2$ & $14.80 \pm 1.60$ & $0.678 \pm 0.032$ & $0.1100 \pm 0.0330$ \\
N2N-iMF & $\mathbf{14.02 \pm 2.38}$ & $\mathbf{0.731 \pm 0.051}$ & $\mathbf{0.0462 \pm 0.0123}$ \\
\bottomrule
\end{tabular}
\endgroup
\end{table}

\begin{figure}[H]
\centering
\includegraphics[width=\linewidth]{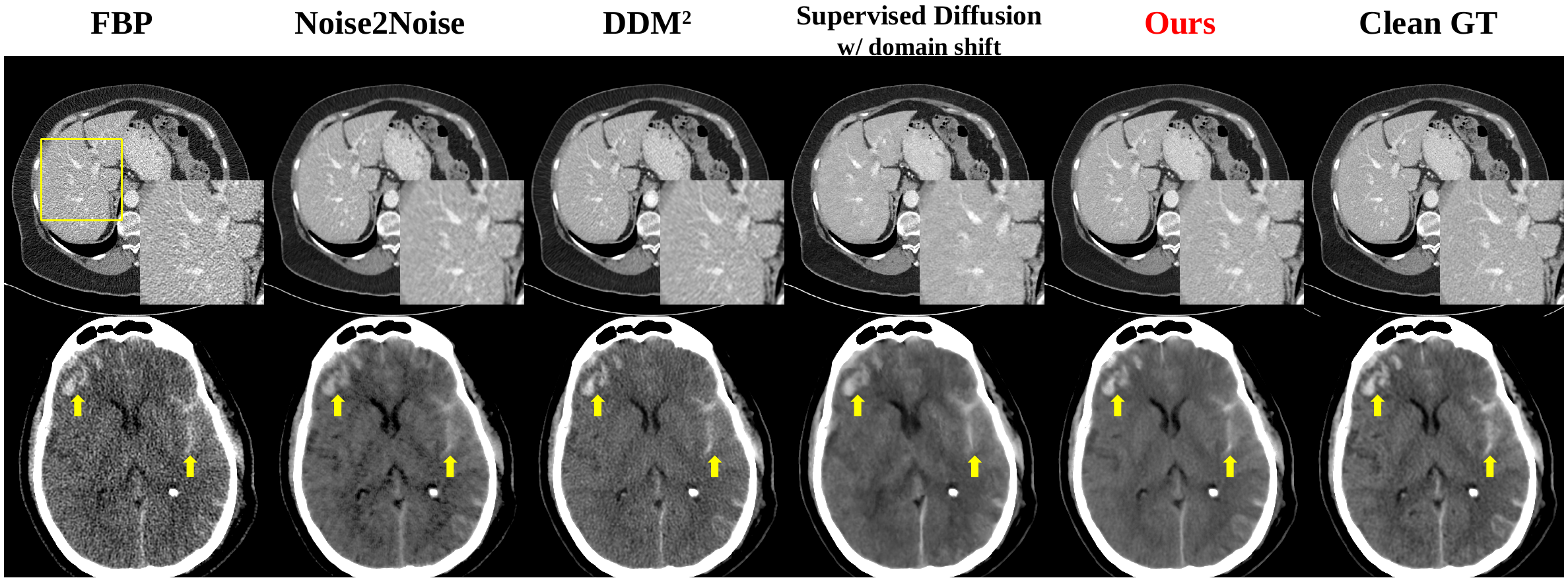}
\caption{Denoising results on simulated abdominal CT (top) and brain CT (bottom). N2N-iMF preserves small vessels in the magnified abdominal region as well as the brain structures indicated by yellow arrows.}
\label{fig:simulated}
\end{figure}

Qualitative examples in Fig.~\ref{fig:simulated} reflect these quantitative gains. In the magnified abdominal inset, N2N-iMF delineates small vessels more clearly than Noise2Noise, whereas the supervised diffusion model trained under mismatched noise and reconstruction conditions exhibits altered local texture. In the brain CT example, N2N-iMF suppresses background noise while keeping the structures marked by yellow arrows clearly visible.

\FloatBarrier
\subsection{Ablation studies}
\subsubsection{Comparison of expedited sampling methods}
At $N_{\mathrm{FE}}=3$ and $K=10$, N2N-iMF ranked first among the four sampling methods across every mean metric on both simulated datasets (Table~\ref{tab:expedited}). Compared with DDIM and EDM, these improvements were statistically significant for all three metrics ($p<0.001$ on brain CT and $p<0.01$ on abdominal CT, two-sided $t$-tests).

\begin{table}[H]
\centering
\singlespacing
\caption{Sampling comparison at $N_{\mathrm{FE}}=3$ and $K=10$ on brain CT ($n=16$) and abdominal CT ($n=3$). Values are mean $\pm$ SD; bold indicates the best mean in each dataset.}
\label{tab:expedited}
\vspace{6pt}
\begingroup
\fontsize{9}{11}\selectfont
\setlength{\tabcolsep}{4pt}
\renewcommand{\arraystretch}{1.18}
\begin{tabular}{llccc}
\toprule
Family & Method & MAE (HU) $\downarrow$ & SSIM $\uparrow$ & LPIPS $\downarrow$ \\
\midrule
\multicolumn{5}{l}{\textit{Brain CT}} \\
Diffusion & DDIM & $7.68 \pm 3.21$ & $0.520 \pm 0.098$ & $0.1905 \pm 0.0598$ \\
 & EDM & $3.41 \pm 0.41$ & $0.718 \pm 0.041$ & $0.0857 \pm 0.0165$ \\
Flow & Flow matching & $2.75 \pm 0.37$ & $0.748 \pm 0.052$ & $0.0639 \pm 0.0199$ \\
 & N2N-iMF & $\mathbf{2.60 \pm 0.36}$ & $\mathbf{0.766 \pm 0.035}$ & $\mathbf{0.0497 \pm 0.0131}$ \\
\midrule
\multicolumn{5}{l}{\textit{Abdominal CT}} \\
Diffusion & DDIM & $19.37 \pm 1.82$ & $0.648 \pm 0.039$ & $0.0953 \pm 0.0065$ \\
 & EDM & $29.33 \pm 1.93$ & $0.297 \pm 0.013$ & $0.1825 \pm 0.0311$ \\
Flow & Flow matching & $14.47 \pm 2.40$ & $0.724 \pm 0.049$ & $0.0539 \pm 0.0136$ \\
 & N2N-iMF & $\mathbf{14.02 \pm 2.38}$ & $\mathbf{0.731 \pm 0.051}$ & $\mathbf{0.0462 \pm 0.0123}$ \\
\bottomrule
\end{tabular}
\endgroup
\end{table}

\begin{figure}[H]
\centering
\includegraphics[width=\linewidth]{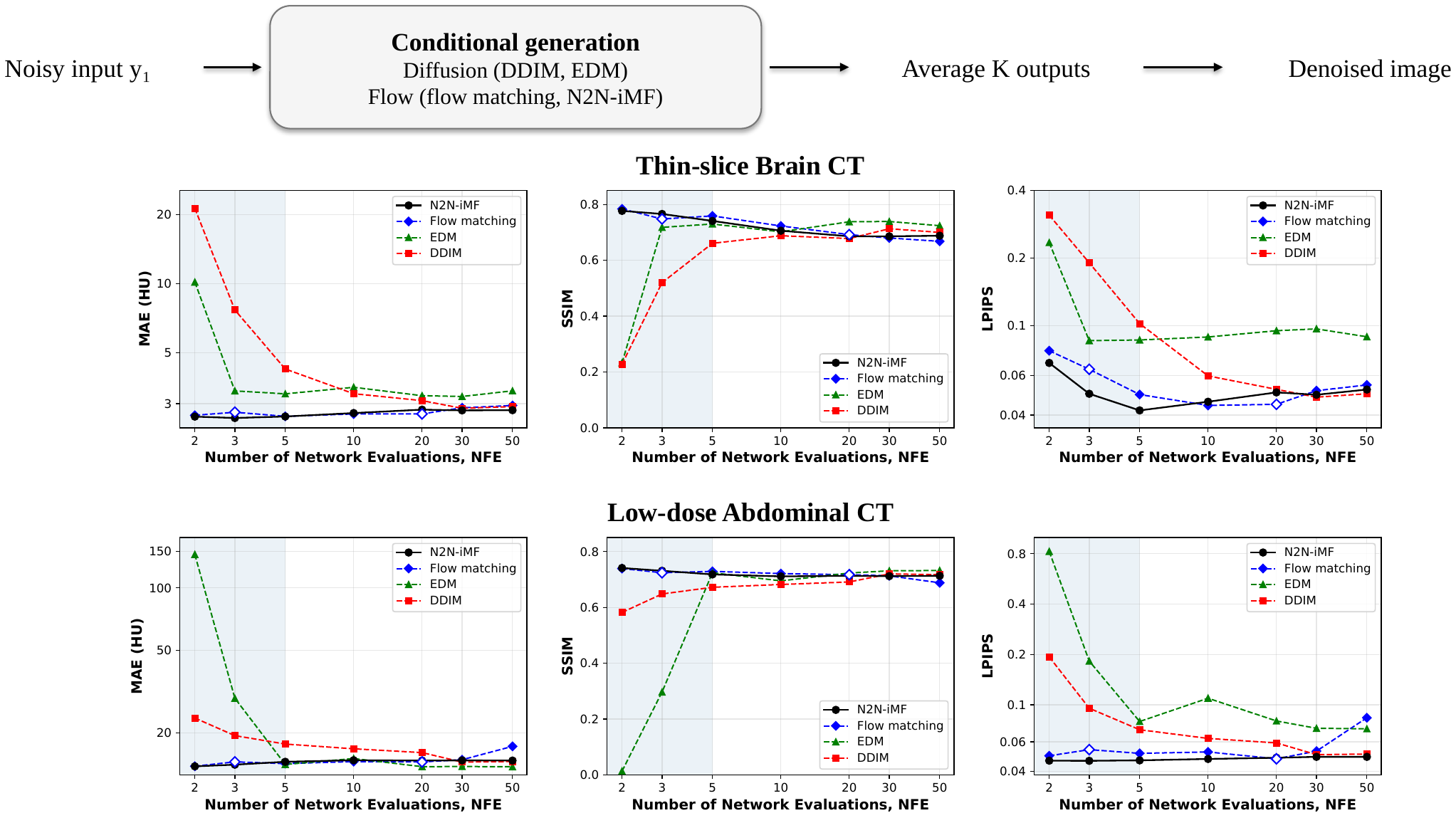}
\caption{Conditional generative denoising framework and sampling comparison at $K=10$. Shading marks the few-step regime ($N_{\mathrm{FE}}\leq5$).}
\label{fig:nfe}
\end{figure}

Even with three evaluations per output, N2N-iMF achieved a lower MAE than 50-evaluation DDIM on both brain CT (2.60 vs.\ 2.91~HU) and abdominal CT (14.02 vs.\ 14.53~HU; Fig.~\ref{fig:nfe}). Flow matching likewise maintained low MAE and high SSIM at two or three evaluations. In contrast, DDIM improved gradually as $N_{\mathrm{FE}}$ increased, while EDM showed a sharp drop in abdominal MAE between three and five evaluations before improving further up to $N_{\mathrm{FE}}=50$.

\FloatBarrier
\subsubsection{Effects of \texorpdfstring{$N_{\mathrm{FE}}$}{NFE} and \texorpdfstring{$K$}{K}}
For N2N-iMF at $K=10$, two or three evaluations yielded the lowest abdominal MAE and highest SSIM, while LPIPS varied little across the tested $N_{\mathrm{FE}}$ range (Fig.~\ref{fig:nfe}). On brain CT, increasing $N_{\mathrm{FE}}$ from three to five improved LPIPS but lowered SSIM.

\begin{figure}[H]
\centering
\includegraphics[width=\linewidth]{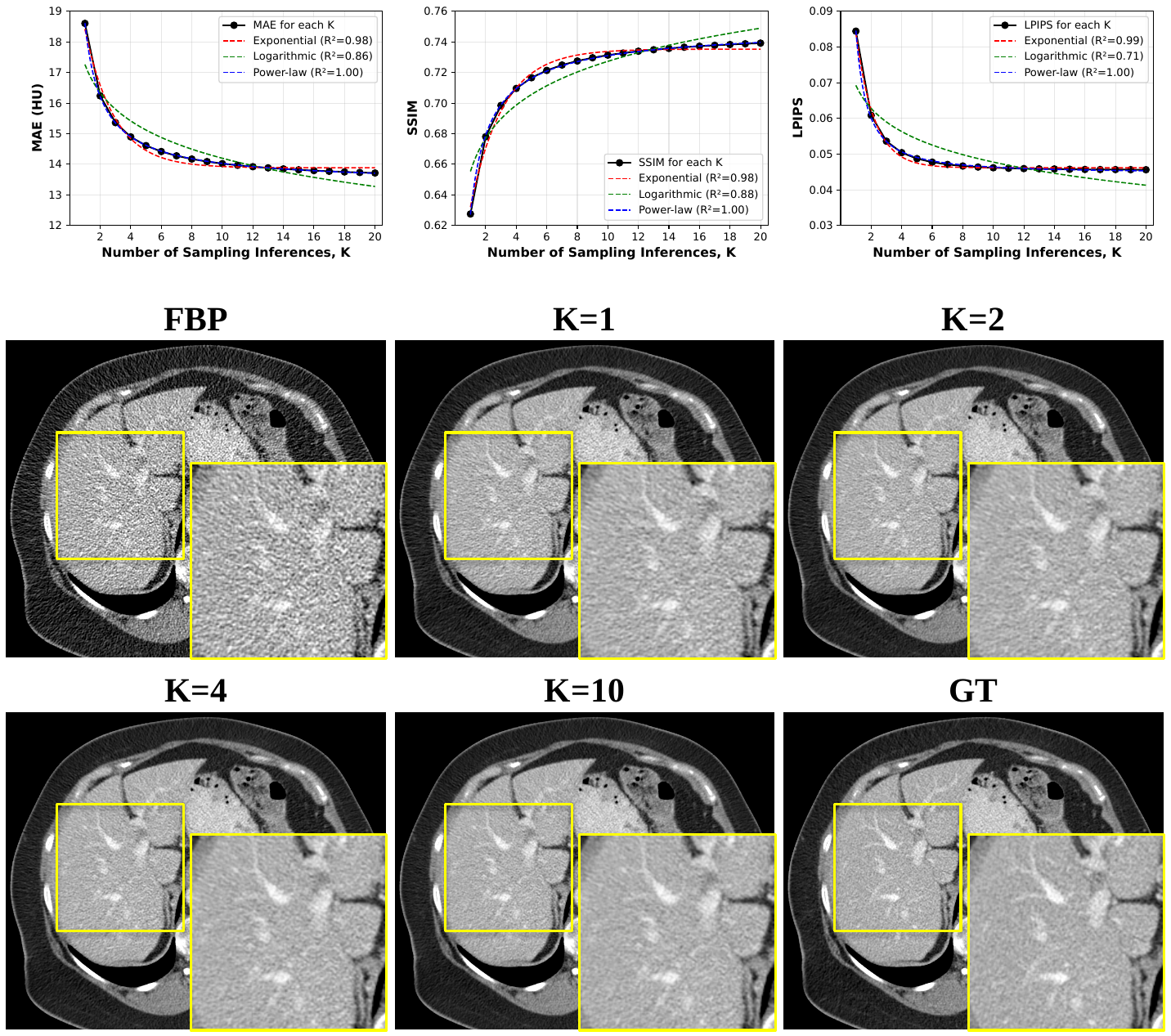}
\caption{Effect of $K$ at $N_{\mathrm{FE}}=3$ on abdominal CT ($n=3$). Points denote means across the three test cases, and dashed lines indicate fitted curves. Representative slices compare FBP, $K=1,2,4,10$, and ground truth (GT), with yellow boxes marking the magnified regions.}
\label{fig:k}
\end{figure}

At $N_{\mathrm{FE}}=3$, increasing $K$ from 1 to 20 reduced abdominal MAE from 18.60 to 13.72~HU, increased SSIM from 0.628 to 0.739, and reduced LPIPS from 0.0844 to 0.0456 (Fig.~\ref{fig:k}). A similar pattern held on brain CT (prior to adversarial fine-tuning), where MAE decreased from 3.60 to 2.58~HU, SSIM increased from 0.616 to 0.782, and LPIPS declined from 0.1074 to 0.0610. On both datasets, gains were smaller at higher $K$.

\subsubsection{Effect of adversarial fine-tuning}
At $N_{\mathrm{FE}}=3$ and $K=10$, PatchGAN adversarial fine-tuning reduced brain CT LPIPS by 22\% (from 0.0637 to 0.0497), while slightly lowering MAE from 2.65 to 2.60~HU and changing SSIM from 0.769 to 0.766.

\clearpage
\subsection{Performance on real-world PCCT}
\begin{figure}[H]
\centering
\includegraphics[width=\linewidth]{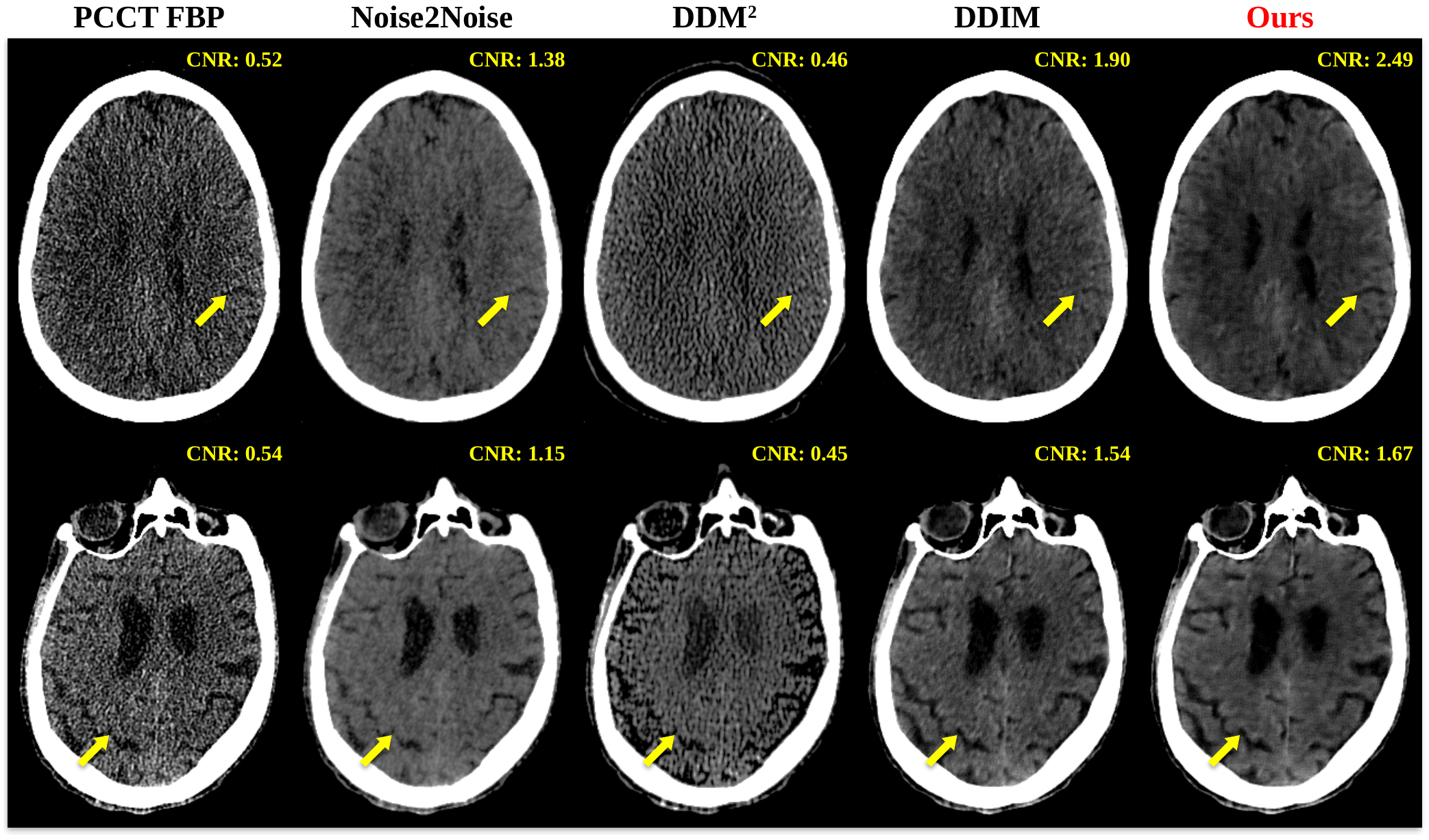}
\par\vspace{8pt}
\includegraphics[width=0.76\linewidth]{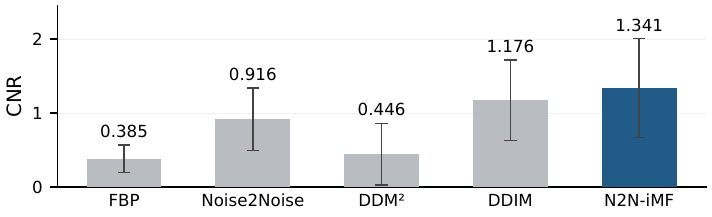}
\caption{Denoising results on real-world thin-slice brain PCCT. Yellow arrows highlight cortical sulci, and the bar chart summarizes mean gray--white matter CNR ($\pm$ SD) across the eight test cases.}
\label{fig:pcct}
\end{figure}

Across the eight PCCT test cases, N2N-iMF achieved the highest mean gray--white matter CNR of 1.341, compared with 0.916 for Noise2Noise and 1.176 for DDIM (Fig.~\ref{fig:pcct}). Visually, it suppressed background noise while maintaining clear delineation of the cortical sulci marked by yellow arrows.

\FloatBarrier
\section{Discussion}
\label{sec:discussion}

In this study, we developed N2N-iMF to accelerate self-supervised generative CT denoising without sacrificing image quality. Trained directly on paired noisy observations, the model generates independent realizations whose average suppresses noise while preserving anatomical detail. Using projection splitting for simulated abdominal CT and adjacent-slice pairing for simulated brain CT, N2N-iMF outperformed the compared self-supervised methods in both image fidelity and perceptual similarity (Table~\ref{tab:main}). On real-world PCCT scans, it likewise achieved the highest mean gray--white matter CNR while preserving the cortical sulci highlighted in Fig.~\ref{fig:pcct}.

Learning from noisy pairs eliminates the need for both clean reference targets and an explicit CT noise model, allowing the generator to adapt directly to the noise characteristics of the target acquisition setting. By contrast, the supervised model trained under mismatched noise and reconstruction conditions altered local texture and small structures (Fig.~\ref{fig:simulated}). N2N-iMF also bypasses the preliminary noise-estimation step that DDM$^2$ relies on to match an input to an intermediate diffusion state~\cite{xiang2023ddm2}; inaccuracies in that estimate may explain the residual noise seen in DDM$^2$'s PCCT outputs (Fig.~\ref{fig:pcct}). In addition, because denoising is performed by averaging generated realizations, the sample count $K$ can be adjusted at inference to control denoising strength and computation without retraining.

Few-step generation is what makes this adjustable averaging computationally practical. Generating a single output took 0.43~s with N2N-iMF versus 7.29~s for 50-evaluation DDIM, markedly lowering the cost of repeated sampling without distillation. At $N_{\mathrm{FE}}=3$ and $K=10$, N2N-iMF significantly outperformed DDIM and EDM across all three metrics ($p<0.001$ on brain CT and $p<0.01$ on abdominal CT, two-sided $t$-tests; Table~\ref{tab:expedited}), while mean MAE, SSIM, and LPIPS also favored N2N-iMF over flow matching. Numerical discretization error from large sampling steps may contribute to the higher errors observed for DDIM and EDM at three evaluations~\cite{song2021ddim,karras2022edm}. By comparison, straight conditional paths in flow matching simplify velocity regression~\cite{lipman2023fm}, while iMF directly learns the velocity averaged over a time interval to support larger updates~\cite{geng2025meanflow,geng2026imf}. For N2N-iMF at $K=10$, increasing $N_{\mathrm{FE}}$ from three to five improved brain CT LPIPS but reduced SSIM (Fig.~\ref{fig:nfe}).

\subsection{Limitations}
Several limitations of this study should be noted. First, the real-world PCCT evaluation covered only eight test cases without low-noise reference images, and thus relied on CNR and visual inspection without assessing diagnostic accuracy or false-positive and false-negative findings. Second, the adjacent-slice pairing strategy used for brain CT and PCCT assumes closely matching anatomy between neighboring slices. Future work will expand evaluation to larger clinical datasets, include paired low-noise references where available, and conduct reader studies to evaluate anatomical fidelity and lesion detection accuracy.

\section{Conclusion}
We have presented N2N-iMF, a self-supervised CT denoising method that combines few-step generative modeling with sample averaging. Without requiring clean reference images or model distillation, a single trained model supports adjustable averaging counts at inference. Using three evaluations per output, N2N-iMF achieved lower MAE than 50-evaluation DDIM on both simulated datasets while reducing per-output generation time from 7.29 to 0.43~s, and delivered higher gray--white matter CNR than Noise2Noise and DDIM on real-world PCCT.

\bibliographystyle{elsarticle-num}
\bibliography{refs}

\end{document}